\documentclass{article}

\usepackage{arxiv}

\usepackage[utf8]{inputenc} % allow utf-8 input
\usepackage[T1]{fontenc}    % use 8-bit T1 fonts
\usepackage{hyperref}       % hyperlinks
\usepackage{url}            % simple URL typesetting
\usepackage{booktabs}       % professional-quality tables
\usepackage{amsfonts}       % blackboard math symbols
\usepackage{nicefrac}       % compact symbols for 1/2, etc.
\usepackage{microtype}      % microtypography
\usepackage{lipsum}		% Can be removed after putting your text content
\usepackage{graphicx}
\usepackage{natbib}
\usepackage{doi}
\usepackage{amsmath}
\usepackage{multirow}
\usepackage{amsmath,amssymb}      % 数学符号: \emptyset, \nu, \zeta, \min 等
\usepackage{algorithm}            % 提供 \begin{algorithm}
\usepackage[noend]{algpseudocode} % 提供 \begin{algorithmic} 和伪代码命令
\algrenewcommand\algorithmiccomment[1]{\hfill // #1}
\usepackage{array}
\title{RexDrug: Reliable Multi-Drug Combination Extraction through Reasoning-Enhanced LLMs}

\author{
Zhijun Wang$^{1}$,
Ling Luo$^{1,*}$,
Dinghao Pan$^{1}$,
Huan Zhuang$^{2}$,
Lejing Yu$^{2}$,
Yuanyuan Sun$^{1}$,
Hongfei Lin$^{1}$\\
\\
$^{1}$School of Computer Science and Technology, Dalian University of Technology,\\
No.2 Linggong Road, Ganjingzi District, Liaoning 116024, China\\
\\
$^{2}$Cancer Hospital of Dalian University of Technology, Liaoning Cancer Hospital \& Institute,\\
No.44 Xiaoheyan Road, Dadong District, Liaoning 110042, China\\
\\
$^{*}$Corresponding author: lingluo@dlut.dlut.edu.cn
}
\date{}
\renewcommand{\headeright}{}
\renewcommand{\undertitle}{}
\renewcommand{\shorttitle}{}
\hypersetup{
pdftitle={RexDrug: Reliable Multi-Drug Combination Extraction through Reasoning-Enhanced LLMs},
pdfsubject={BioNLP},
pdfauthor={Zhijun Wang, Ling Luo, Dinghao Pan, Huan Zhuang, Lejing Yu, Yuanyuan Sun, Hongfei Lin},
pdfkeywords={Drug Combination Extraction, N-ary Relation Extraction, Large Language Model, Explainable Reasoning},
}

\begin{document}
\maketitle

\begingroup
\renewcommand{\thefootnote}{}
\footnotetext{\textcopyright{} 2026 IEEE. Personal use of this material is permitted. Permission from IEEE must be obtained for all other uses, in any current or future media, including reprinting/republishing this material for advertising or promotional purposes, creating new collective works, for resale or redistribution to servers or lists, or reuse of any copyrighted component of this work in other works.}
\addtocounter{footnote}{-1}
\endgroup

\begin{abstract}
Automated Drug Combination Extraction (DCE) from large-scale biomedical literature is crucial for advancing precision medicine and pharmacological research. However, existing relation extraction methods primarily focus on binary interactions and struggle to model variable-length \textit{n}-ary drug combinations, where complex compatibility logic and distributed evidence need to be considered. To address these limitations, we propose RexDrug, an end-to-end reasoning-enhanced relation extraction framework for \textit{n}-ary drug combination extraction based on large language models. RexDrug adopts a two-stage training strategy. First, a multi-agent collaborative mechanism is utilized to automatically generate high-quality reasoning traces for supervised fine-tuning. Second, reinforcement learning with a multi-dimensional reward function specifically tailored for DCE is applied to further refine reasoning quality and extraction accuracy. Extensive computational evaluations on the DrugComb dataset show that RexDrug consistently outperforms state-of-the-art baselines for \textit{n}-ary extraction. Additional evaluation on the DDI13 corpus confirms its generalizability to binary drug–drug interaction tasks. Human evaluation and automatic reasoning metrics further indicate that RexDrug produces coherent medical reasoning while accurately identifying complex therapeutic regimens. These results establish RexDrug as a scalable and reliable solution for complex biomedical relation extraction from unstructured text.
\end{abstract}

% keywords can be removed

\section{Introduction}
Drug combination therapy has become the standard treatment for complex diseases such as cancer, HIV/AIDS, malaria, and tuberculosis due to its significant synergistic effects \cite{B2,D1,B4,E3}. However, the space of potential drug combinations is enormous, and many candidate combinations may offer limited benefit or even induce severe adverse reactions \cite{zhang2015mixture, wu2022machine}. Given the rapid growth of biomedical literature, the accurate and efficient identification of safe and effective drug combinations from large-scale unstructured text has become a critical imperative for evidence-based medicine and complex treatment planning \cite{lopez2017combine, luo2022biored}.

To address this need, Tiktinsky et al. \cite{tiktinsky2022dataset} introduced the task of Drug Combination Extraction (DCE) and constructed DrugComb, the first dataset supporting variable-length \textit{n}-ary drug combination relation extraction. Unlike traditional binary relation extraction, the DCE task requires the identification of an arbitrary number of drug entities and their associated therapeutic effects within real-world clinical scenarios. This task is complicated by the fact that evidence for such combinations is often distributed across multiple sentences, making the modeling of long-range semantic dependencies and flexible combination structures particularly difficult.

Existing approaches for this task generally fall into three paradigms. First, pipeline-based methods \cite{zaikis2021tp,kilicoglu2020broad,kang2017eliie} typically perform named entity recognition (NER) followed by relation classification. Although modular and easy to implement, this design ignores the interdependency between subtasks, leading to error propagation where NER failures directly compromise classification performance \cite{zhang2020deep}. Some DCE studies further simplify the task by assuming gold-standard entity annotations and focusing solely on relation classification \cite{zhang2022cnn, shi2024subge, zhang2025reading}. While this setting facilitates controlled evaluation, it limits applicability in real-world scenarios where entities are unknown. Second, end-to-end methods have shown promising results by jointly modeling entities and relations \cite{luo2020neural,zuo2022span,lai2021joint}. However, they are primarily optimized for fixed binary structures, struggling to capture the complex semantics inherent in \textit{n}-ary relations. Third, large language models (LLMs) have recently shown strong potential for generative information extraction \cite{yang2025rise, zhang2023aligning, wang2023instructuie}. Their ability to produce structured outputs makes them appealing for complex relation extraction tasks. Nevertheless, in biomedical settings, LLMs are typically used for direct answer generation without explicit reasoning supervision \cite{luo2024taiyi,dagdelen2024structured}. This can easily produce hallucinated outputs, impairing the interpretability and credibility of the extraction results. 

To address these limitations, we introduce RexDrug, a unified end-to-end generative framework for \textit{n}-ary DCE inspired by the expert analytical paradigm of \emph{evidence retrieval, causal induction, and evidence synthesis}. RexDrug is designed to endow LLMs with systematic pharmacological reasoning capabilities, organizing source evidence into a structured, extraction-oriented reasoning process. The framework employs a two-stage training strategy. In the first stage, a multi-agent collaborative mechanism automatically synthesizes reasoning traces, which are utilized for supervised fine-tuning (SFT) to establish preliminary reasoning strategies and a structured output format. In the second stage, we implement reinforcement learning (RL) with a multi-dimensional reward function tailored to drug combination identification. This function incorporates metrics for reasoning structure compliance, matching accuracy, and relation type prediction to refine the quality and interpretability of the generated output.

Extensive computational evaluations on the DrugComb dataset \cite{tiktinsky2022dataset} show that RexDrug outperforms existing state-of-the-art baselines in \textit{n}-ary relation extraction. Furthermore, evaluation on the DDI13 \cite{herrero2013ddi} benchmark confirms its generalizability to binary drug relation extraction. Human evaluation indicates that RexDrug produces coherent, hierarchical reasoning traces that provide traceable evidentiary support for extracted pharmacological relations.

In summary, our primary contributions are as follows:
\begin{itemize}
    \item We propose RexDrug, a reasoning-augmented framework that reformulates \textit{n}-ary drug-combination extraction as an interpretable generative paradigm. This approach achieves high performance while empowering the model with expert-like, regimen-aware reasoning capabilities.
    \item We design an automated multi-agent collaborative mechanism to synthesize high-quality reasoning traces. This effectively addresses the scarcity of annotated pharmacological logic. Furthermore, we demonstrate the efficacy of multi-dimensional reward functions in capturing the nuances of complex therapeutic regimens.
    \item We demonstrate that RexDrug yields substantial performance gains on both \textit{n}-ary and binary datasets, and further show via rigorous human evaluation that its reasoning trajectories are more coherent and evidence-aligned.
\end{itemize}

% This is an example of a new parapgraph with a numbered footnote\footnote{\url{https://data.gov.uk/}} and a second footnote marker.\footnote{Example of footnote text.}

\section{Related Work}
\subsection{\textit{N}-ary Relation Extraction in Biomedicine} 
Biomedical relation extraction has progressed from identifying intra-sentence binary interactions to capturing complex multi-entity associations \cite{lee2020biobert,gu2021domain}. In physiological and pharmacological contexts, such as enzyme-substrate-inhibitor mechanisms and synergistic multi-drug regimens, \textit{n}-ary relations often capture critical dependencies among multiple entities. Early approaches relied on neural architectures to model cross-sentence context and inter-entity dependencies \cite{peng2017cross,song2018n}. With the advent of Transformer architectures, subsequent work has combined graph representations with attention mechanisms \cite{jia2019document,zhao2020incorporating} to better capture long-range semantic dependencies.

However, whether using pipeline architectures \cite{zhang2025reading} or end-to-end frameworks \cite{jiang2023end}, existing efforts have largely focused on improving extraction accuracy. Yet pharmacological relation determination critically depends on logical consistency. Current methods often lack biologically grounded interpretability, making them insufficient for rigorous scientific evidence review.

\subsection{Reasoning Enhancement Strategies in Biomedicine} 

To improve both interpretability and robustness in biomedical information extraction, recent studies have explored reasoning-augmented strategies that decompose the task into intermediate steps \cite{bian2023inspire,liu2024era}. Meanwhile, the strong generative capabilities of LLMs have gained increasing attention. When combined with RL, LLMs can be guided to generate task-specific reasoning using carefully designed reward signals, facilitating capabilities such as causal reasoning \cite{li2024rt} and data selection \cite{qiu2025open}. 

Nevertheless, most existing efforts remain confined to named entity recognition or simple binary relation extraction, failing to effectively address the higher-order logical challenges inherent in \textit{n}-ary relation extraction. Moreover, the scarcity of high-quality, structured pharmacological reasoning traces constitutes a major technical bottleneck in this domain. To address these challenges, we propose a multi-agent collaborative data generation framework coupled with a reward-guided training paradigm tailored for \textit{n}-ary tasks. RexDrug not only targets extraction accuracy, but also aims to produce reasoning traces with medical analysis capability, thereby bridging the gap between automated extraction and expert logical verification.

\section{Task Formulation}
Drug Combination Extraction (DCE) aims to identify pharmacological relationships involving a variable number of drugs from unstructured biomedical literature. Unlike traditional approaches that treat this solely as a relation classification problem \cite{tiktinsky2022dataset}, we reformulate the task as a generative process driven by explicit reasoning. Formally, given a target sentence $S$ and its surrounding context $C$, let $E(S)=\{e_1,...,e_m\}$ denote the set of drug entities appearing in $S$. Let $Y$ be the predefined set of drug-combination effect types, including POS (positive effect) and OTHER (a combination exists, but the effect is either negative or undetermined). Conditioned on the input $X=(S,C)$, the model leverages the contextual background to identify all drug combinations $P_i$ mentioned in the target sentence $S$, where $P_i \subseteq E(S)$ and $|P_i|\geq2$. The model then assigns a label $y_i\in Y$ to each $P_i$. In addition, it generates a pharmacological reasoning trace $H$, which integrates evidence from the surrounding context and performs inference to justify the extracted results. The final output is formalized as $R(X)=(H,\{(P_i,y_i)\}_{i=1}^I)$, where $I$ denotes the number of predicted drug combinations for the given input $X$. If no drug combination is mentioned in $S$, we output the label NO\_COMB with an empty set of combinations (i.e., $I=0$ and $\{(P_i,y_i)\}_{i=1}^I=\emptyset$).

\begin{figure*}[t]
\centering
    \includegraphics[width=\textwidth]{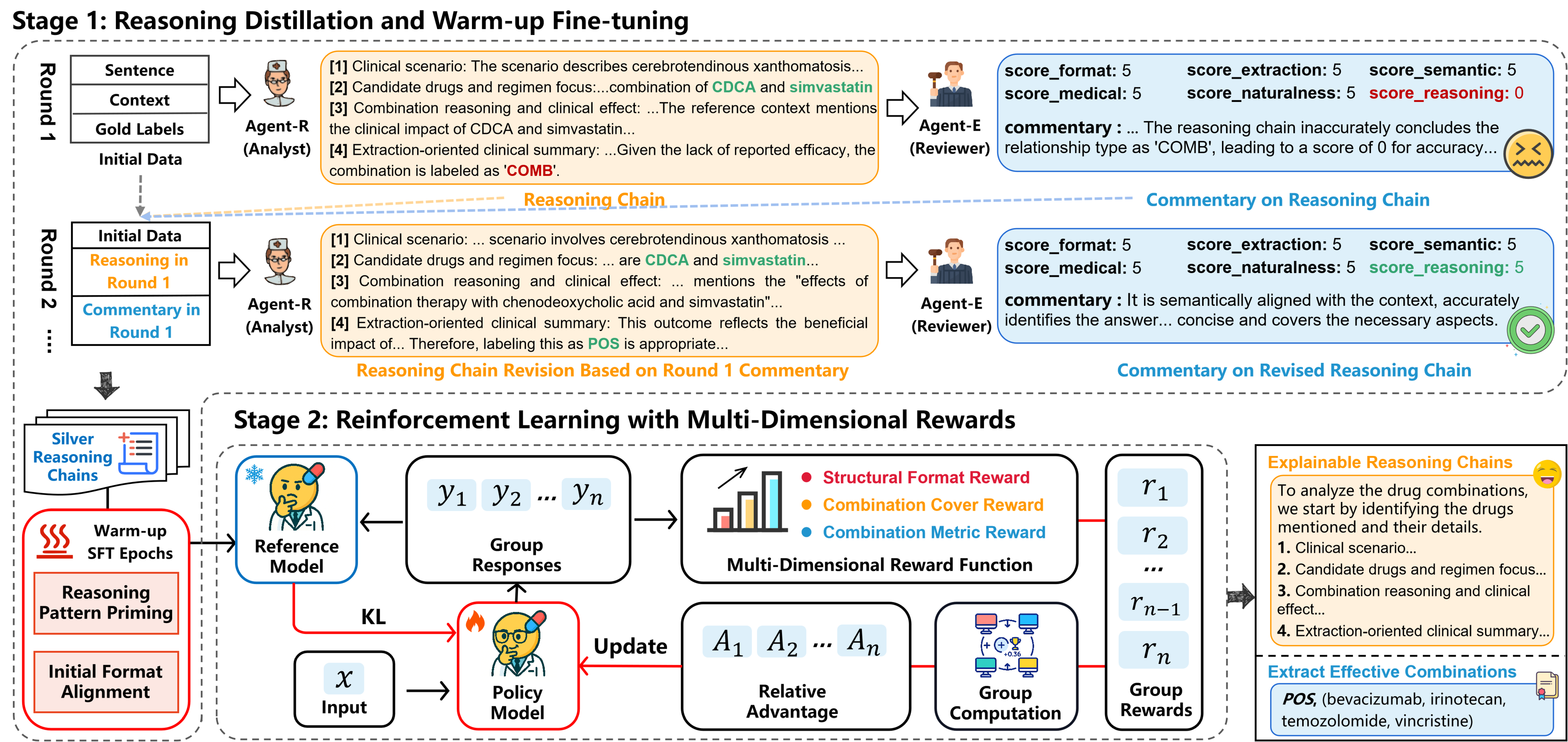}
    \caption{Overall framework of RexDrug.}
    \label{Fig2}
\end{figure*}

\section{Method}
Based on the above task formulation, the overall architecture of RexDrug is illustrated in Fig.~\ref{Fig2}. It is designed to endow an LLM with expert-like biomedical reasoning through two progressive stages: (1) Warm-up Fine-tuning with Multi-Agent Reasoning Distillation, where we employ a multi-agent collaboration strategy to generate high-quality reasoning traces that emulate expert thought processes to initialize the model. (2) Reinforcement Learning with Multi-Dimensional Rewards, where we introduce multi-dimensional reward functions tailored to the DCE task and apply the Group Relative Policy Optimization (GRPO) \cite{shao2024deepseekmath} algorithm to further optimize the model under reward guidance.

\begin{table*}[t]
\centering
\caption{Instruction used for relation extraction.}
\label{tab:instruction}
\begin{tabular}{p{0.95\textwidth}}
\hline
Identify all possible drug combination relationships mentioned in the target sentence, and determine their combined usage effect category. You may refer to the surrounding paragraph for contextual reasoning. Then provide your reasoning and final answer in two parts: (1) First, output your reasoning inside \texttt{<think>...</think>}. Inside \texttt{<think>}, write four numbered sections: [1] Overview of the target document context; [2] Candidate drugs; [3] Combination reasoning and treatment or experimental effects; [4] Extraction-oriented medical summary. Under each section, use bullet points starting with \texttt{- } with short, clinically oriented sentences. Keep the total reasoning concise (about 100--200 words). (2) Immediately after \texttt{</think>}, output ONLY the final relation extraction result inside an \texttt{<answer>} tag. Inside \texttt{<answer>...</answer>}, return a valid JSON array. \\
\hline
\end{tabular}
\end{table*}

\subsection{Stage 1: Warm-up Fine-tuning with Multi-Agent Reasoning Distillation}

To address the scarcity of annotated pharmacological reasoning, RexDrug employs two LLM agents as a Medical Reasoning Analyst and a Medical Expert Reviewer. The Analyst generates step-by-step reasoning conditioned on the original text and the human-annotated tuple labels, following the prompt template in Table \ref{tab:instruction}. Providing human-annotated labels serves as an anchor to constrain the Analyst’s reasoning direction and encourage alignment with the annotated facts. However, even with explicit instructions, the Analyst may still produce factually inconsistent reasoning traces due to overconfidence. To mitigate this issue, the Reviewer evaluates each reasoning trace according to six stringent criteria: Format Compliance, Extraction Consistency, Semantic Faithfulness, Medical Validity, Narrative Naturalness, and Reasoning Quality. Each criterion is scored on a scale from 0 to 5. A reasoning trace is accepted only if all scores are at least 4; otherwise, the Reviewer provides concise feedback to guide the Analyst in revising the chain.

The complete Analyst and Reviewer prompts are available in the code repository. Each instance undergoes up to three generation-review-feedback rounds and is discarded if no acceptable trace is obtained. On DrugComb, this procedure retains 1098 of 1362 training instances (80.6\%). Supplementary Section S1 examines label conditioning through a label-blind rejection-sampling control, and Supplementary Section S2 audits the 264 excluded instances for potential selection bias.

After obtaining high-quality pharmacological reasoning traces, we perform SFT on the base model. This warm-up fine-tuning phase is primarily designed to instill fundamental pharmacological knowledge and enforce a structured reasoning format. Consequently, this integrated process not only establishes a stable policy space for subsequent reinforcement learning, but also effectively mitigates common issues in large-scale \textit{n}-ary extraction, such as format collapse and logical discontinuities \cite{chu2025sft}. 

\subsection{Stage 2: Reinforcement Learning with Multi-Dimensional Rewards}
The SFT model is further optimized using GRPO on training pairs $(X,\mathcal{G})$, where $X$ is the input text and $\mathcal{G}$ is the set of gold drug combinations. The objective of this stage is to encourage the model to generate extraction results and reasoning traces that are not only structurally valid but also pharmacologically accurate and logically coherent (Fig.~\ref{Fig2} (Stage 2)). For each input, the policy $\pi_\theta$ generates a group of responses $\{o_i\}_{i=1}^{K}$, whose rewards are normalized within the group to estimate their relative advantages. The frozen SFT checkpoint serves as the reference policy $\pi_{\theta_{\mathrm{ref}}}$, and Kullback-Leibler (KL) divergence regularization limits deviation from this policy during optimization. For the DCE task, we design three types of targeted reward signals $r_i$ to calculate the relative advantage $A_i$ within the response group:

(1) Structural Format Reward: This reward is designed to ensure that the model-generated reasoning traces conform to a standardized output format:
\begin{equation}
r_{\mathrm{format}}(o_i) =
\begin{cases}
0, & \text{if } I_{tag} = \text{False} \\
0.5 + s_{\mathrm{t}}(o_i) + s_{\mathrm{a}}(o_i), & \text{if } I_{tag} = \text{True}
\end{cases}
\end{equation}
Here, $I_{tag}$ verifies whether the output contains complete \texttt{<think>} and \texttt{<answer>} tags; the cognitive-path alignment score $s_t(o_i)$ evaluates whether the reasoning in \texttt{<think>} follows the required four section structure, and assigns lower scores when any required step is missing; the extraction-format score $s_a(o_i)$ ensures that the \texttt{<answer>} content is a valid, machine-readable JSON list.

(2) Combination Coverage Reward: This reward introduces a coverage metric that decomposes the DCE task into a learnable sub-goal, thereby addressing the issue of reward sparsity. Additionally, considering the high proportion of negative examples (NO\_COMB), a model that consistently outputs empty predictions might receive undeserved high scores. To prevent this behavior, we introduce a penalty for incorrect empty predictions. Let $\mathcal{P}=\{P_i\}_{i=1}^I$ denote the predicted set of combinations. By calculating the maximum coverage between each predicted combination $P_i\in \mathcal{P}$ and all ground-truth combinations $g_j\in \mathcal{G}$, the final drug combination reward integrates both the coverage reward and the penalty term.

\begin{equation}
\begin{aligned}
r_{\mathrm{comb\_cover}}
&=\frac{1}{\max(1,|\mathcal{P}|)}
\sum_{P_i\in\mathcal{P}}
\max_{g_j\in\mathcal{G}}
\frac{|P_i\cap g_j|}{|g_j|} \\
&\quad
-\mathbb{I}\!\left[
\mathcal{P}=\emptyset \land \mathcal{G}\neq\emptyset
\right].
\end{aligned}
\label{eq5}
\end{equation}

(3) Drug Combination Metric Reward: To improve extraction accuracy without compromising interpretability, we design a metric-based reward using commonly adopted DCE evaluation metrics, namely Exact and Partial F1 \cite{tiktinsky2022dataset}. Exact Match requires the relation type and the predicted drug entity set to be fully identical to the ground truth, whereas Partial Match encourages the model to capture subset relations within higher-order combinations. The final metric reward combines Exact F1 and Partial F1, assigning a slightly higher weight to Exact F1 to encourage fully correct predictions:

\begin{equation}
    r_{\mathrm{comb\_metric}}=\frac{2}{3}\cdot F1_{\mathrm{exact}}+\frac{1}{3}\cdot F1_{\mathrm{partial}}
\end{equation}

The final reward combines these components with weights $\alpha_1=0.2$, $\alpha_2=0.1$, and $\alpha_3=0.7$. The metric reward receives the largest weight because it directly represents the extraction objective, while the format and coverage rewards provide auxiliary structural and intermediate supervision. Reward-weight sensitivity is reported in Supplementary Section S3.

\begin{equation}
    r_i=\alpha_1\cdot r_{\mathrm{format}}+\alpha_2\cdot r_{\mathrm{comb\_cover}}+\alpha_3\cdot r_{\mathrm{comb\_metric}}
\end{equation}

The corresponding relative advantage score $A_i$ is normalized as follows:

\begin{equation}
    A_i=\frac{r_i-mean(\{r_1,...,r_K\})}{\mathrm{std}(\{r_1,...,r_K\})}
\end{equation}

The GRPO objective balances reward maximization and policy stability. The overall optimization function is defined as:

\begin{equation}
\begin{aligned}
\mathcal{I}_{\mathrm{GRPO}}(\theta)
=&\;
\mathbb{E}_{x\sim\mathcal{D},\,\{o_i\}_{i=1}^{K}\sim\pi_{\theta_{\mathrm{old}}}(\cdot|x)}
\\
&\times
\Bigg[
\frac{1}{K}\sum_{i=1}^{K}
\min\left(\rho_i A_i,
\mathrm{clip}(\rho_i,1-\varepsilon,1+\varepsilon)A_i\right)
\\
&\qquad
-\beta D_{\mathrm{KL}}
\left(\pi_\theta\|\pi_{\theta_{\mathrm{ref}}}\right)
\Bigg].
\end{aligned}
\end{equation}

Here $\rho_i=\frac{\pi_{\theta}(o_i|x)}
{\pi_{\theta_{\mathrm{old}}}(o_i|x)}$ measures the relative policy update ratio, 
and $\varepsilon$ limits the update magnitude through clipping to improve training stability. 
$A_i$ denotes the estimated advantage of the sampled output.

\section{Results}
\subsection{Experimental Datasets and Settings}

We evaluate RexDrug on DrugComb \cite{tiktinsky2022dataset}, which comprises 1634 manually annotated abstracts containing 2 to 15 drug mentions and three relation labels: POS, OTHER, and NO\_COMB. To examine applicability beyond \textit{n}-ary extraction, we apply the same training pipeline to DDI13 \cite{herrero2013ddi}, a binary drug--drug interaction benchmark with four relation types: Mechanism, Effect, Advice, and Int. Int indicates the presence of a DDI without specifying its type. Sentences without an annotated interaction are mapped to NO\_COMB in our unified output format. Dataset statistics are provided in Supplementary Section S3.

Following Tiktinsky et al. \cite{tiktinsky2022dataset}, DrugComb performance is evaluated using exact and partial matching. Exact matching requires complete agreement in relation type and drug membership, whereas partial matching assigns overlap-based credit. Under both schemes, we report Positive Combination F1 for POS relations and Any Combination F1 for POS and OTHER relations. DDI13 is evaluated using precision, recall, and F1 \cite{fei2021span,sun2022mrc4bioer}; a prediction is correct only when both entity boundaries and interaction type match the annotation.

In terms of implementation, we use LLaMA3.1-8B-Instruct and Qwen2.5-7B-Instruct as the base models. We apply low-rank adaptation (LoRA) for fine-tuning in both the SFT and RL stages, and conduct evaluations with five different random seeds, reporting the mean and standard deviation. For reasoning-trace generation on both datasets, GPT-4o
(gpt-4o-2024-11-20) served as the Analyst and GPT-5.1 (gpt-5.1-2025-11-13) as the Reviewer. This asymmetric design follows the common practice of using a stronger model as the evaluator to provide more reliable feedback. For all zero-shot models, we use the same instruction prompts as those used during training, as summarized in Table \ref{tab:instruction}. To ensure reproducibility, detailed hyperparameter settings and seed control are provided in Supplementary Section S3.

\begin{table*}[t]
\centering
\caption{Performance comparison on the DrugComb dataset. \textbf{Bold} indicates the best result and \underline{underline} the second best. Results are reported as mean $\pm$ standard deviation over 5 runs. ``w/o reasoning'' denotes direct extraction without generating a reasoning trace, while ``w/ reasoning'' requires a reasoning trace. DAPT denotes continued domain-adaptive pretraining. ``--'' indicates that the value was not reported. PURE* and RCFIND* are classification baselines that use human-annotated entity mentions, which simplifies the task.}
\label{tab:main_results}

\setlength{\tabcolsep}{5pt}
\resizebox{\textwidth}{!}{%
\begin{tabular}{llcccc}
\toprule
\textbf{Category} & \textbf{Model} &
\textbf{F1(pos,exact)} &
\textbf{F1(pos,partial)} &
\textbf{F1(any,exact)} &
\textbf{F1(any,partial)} \\
\midrule
PURE*   & PubmedBERT+DAPT & 61.8 & 67.7 & 69.4 & 77.5 \\
RCFIND* & PubmedBERT      & 72.0 & 74.9 & 80.3 & 83.3 \\
Seq2Rel & PubmedBERT      & 66.7 & --   & 71.1 & --   \\
\midrule
\multirow{5}{*}{Zero-shot (w/o reasoning)}
& GPT-3.5              & 39.4 & 41.9 & 40.7 & 43.4 \\
& GPT-4o               & 49.2 & 53.6 & 54.5 & 60.6 \\
& llama3.1-8b-instruct & 34.7 & 40.1 & 42.4 & 49.5 \\
& qwen2.5-7b-instruct  & 37.1 & 41.2 & 45.3 & 51.5 \\
& qwen3-30b-instruct   & 46.0 & 51.9 & 51.6 & 59.1 \\
\midrule
\multirow{6}{*}{Zero-shot (w/ reasoning)}
& GPT-3.5              & 27.9 & 31.8 & 29.9 & 31.5 \\
& GPT-4o               & 43.8 & 48.0 & 50.9 & 55.9 \\
& llama3.1-8b-instruct & 30.2 & 38.3 & 36.4 & 44.9 \\
& qwen2.5-7b-instruct  & 29.5 & 34.7 & 34.3 & 40.5 \\
& qwen3-30b-instruct   & 36.2 & 43.2 & 40.1 & 49.0 \\
& qwen3-30b-think      & 39.8 & 44.9 & 43.0 & 47.8 \\
\midrule
\multirow{2}{*}{SFT (w/o reasoning)}
& llama3.1-8b-instruct & $72.9\pm0.4$ & $75.4\pm0.3$ & $80.5\pm0.3$ & $83.4\pm0.3$ \\
& qwen2.5-7b-instruct  & $72.1\pm0.9$ & $74.6\pm0.5$ & $80.0\pm0.6$ & $82.9\pm0.8$ \\
\midrule
\multirow{2}{*}{SFT (w/ reasoning)}
& llama3.1-8b-instruct & $68.9\pm1.2$ & $72.8\pm1.1$ & $75.1\pm0.6$ & $80.3\pm0.5$ \\
& qwen2.5-7b-instruct  & $69.3\pm0.6$ & $72.1\pm0.4$ & $75.2\pm0.7$ & $79.4\pm0.6$ \\
\midrule
\multirow{2}{*}{Only-RL (w/ reasoning)}
& llama3.1-8b-instruct & $40.0\pm1.2$ & $46.2\pm0.9$ & $46.1\pm1.1$ & $53.3\pm0.9$ \\
& qwen2.5-7b-instruct  & $38.1\pm0.9$ & $46.2\pm0.8$ & $43.6\pm1.0$ & $52.6\pm1.3$ \\
\midrule
\multirow{2}{*}{RexDrug (w/ reasoning)}
& llama3.1-8b-instruct
& \textbf{74.6} $\pm$ \textbf{0.7}
& \textbf{77.1} $\pm$ \textbf{0.5}
& \textbf{81.4} $\pm$ \textbf{0.5}
& \textbf{85.3} $\pm$ \textbf{0.6} \\
& qwen2.5-7b-instruct
& \underline{$74.2\pm0.4$}
& \underline{$76.9\pm0.8$}
& \underline{$81.3\pm0.6$}
& \underline{$84.1\pm0.5$} \\
\bottomrule
\end{tabular}%
}
\end{table*}

\begin{table}[t]
\centering
\caption{Auxiliary oracle-entity evaluation on DrugComb. All models are provided with gold drug mentions and predict combination membership and relation type. ``-R'' denotes settings in which the model is required to generate reasoning traces.}
\label{tab:drugcomb_single}

\renewcommand{\arraystretch}{1.08}

\begin{tabular*}{\columnwidth}{@{\extracolsep{\fill}}lcccc@{}}
\toprule
\textbf{Method} & \textbf{Pos-E} & \textbf{Pos-P} & \textbf{Any-E} & \textbf{Any-P} \\
\midrule
PURE*   & 61.8 & 67.7 & 69.4 & 77.5 \\
RCFIND* & 72.0 & 74.9 & 80.3 & 83.3 \\
\midrule
GPT-3.5    & 59.7 & 63.6 & 63.9 & 68.7 \\
GPT-3.5-R  & 44.8 & 49.4 & 48.7 & 53.6 \\
GPT-4o     & 68.0 & 71.0 & 72.2 & 75.9 \\
GPT-4o-R   & 69.1 & 70.6 & 72.1 & 75.6 \\
\midrule
SFT-Qwen   & 78.8$\pm$0.4 & 80.9$\pm$0.6 & 84.7$\pm$0.3 & 86.9$\pm$0.3 \\
SFT-LLaMA  & 80.4$\pm$0.7 & 82.9$\pm$0.9 & 85.2$\pm$0.5 & 87.8$\pm$0.8 \\
\midrule
RexDrug-Qwen
& \textbf{82.4$\pm$0.9}
& \underline{83.9$\pm$0.7}
& \underline{85.4$\pm$0.8}
& \underline{88.9$\pm$0.7} \\
RexDrug-LLaMA
& \underline{82.1$\pm$0.5}
& \textbf{85.9$\pm$0.6}
& \textbf{85.9$\pm$1.2}
& \textbf{90.7$\pm$0.9} \\
\bottomrule
\end{tabular*}

\end{table}

\subsection{Performance on the DrugComb dataset}
To evaluate RexDrug on the DrugComb dataset, we compare it against several representative baselines. PURE \cite{tiktinsky2022dataset} is a PLM-based relation classifier that inserts entity tags into the input to improve entity awareness for drug-combination classification. RCFIND \cite{zhang2025reading} enhances \textit{n}-ary classification via semantic querying and capsule networks. Seq2Rel \cite{jiang2023end} is an end-to-end framework that jointly extracts entities and relations using PubMedBERT with a long short-term memory (LSTM) network.

Notably, PURE* and RCFIND* use human-annotated drug entities and perform relation classification, which is a simpler setting than directly extracting relation tuples. In addition, we conduct extensive computational evaluations under two paradigms, direct structured generation and reasoning-augmented generation, using both open-source LLMs (LLaMA, Qwen) and strong closed-source LLMs (GPT-3.5, GPT-4o). Table \ref{tab:main_results} summarizes the overall performance of all models on the DrugComb:

(1) Even the proprietary GPT-family models exhibit suboptimal performance in the zero-shot setting, suggesting that general-purpose LLMs still struggle with the complexity of biomedical DCE. In contrast, with basic SFT, both LLaMA and Qwen surpass traditional baselines, highlighting the advantage of the generative paradigm for variable-length \textit{n}-ary relation extraction. (2) Explicit reasoning consistently degrades zero-shot performance, indicating that without task-specific training and alignment, requiring the model to generate both explanations and strictly structured outputs increases the burden of generation and constraint adherence, consistent with Nagar et al. \cite{nagar2025llms}. Reasoning-supervised SFT recovers much of this loss but remains below direct SFT, suggesting that it establishes task-specific reasoning and output patterns without fully aligning reasoning generation with extraction accuracy. (3) Only-RL remains substantially inferior to all SFT-based settings, indicating that reward optimization from the base model lacks a stable policy for joint reasoning and extraction. In contrast, initializing GRPO from reasoning-supervised SFT enables RexDrug to outperform both SFT variants. These results demonstrate complementary roles: SFT provides a stable policy for task-specific reasoning and structured generation, while GRPO further aligns it with format compliance, combination coverage, and extraction accuracy. (4) Under the same end-to-end extraction setting, RexDrug improves Pos-Exact F1 over Seq2Rel by approximately 7.9\%. Although PURE* and RCFIND* leverage gold entity annotations, they are still outperformed by RexDrug. For a controlled comparison, Table \ref{tab:drugcomb_single} further reports an auxiliary evaluation in which RexDrug is also provided with gold entity spans. RexDrug not only outperforms both baselines under this setting but also improves upon its own end-to-end results, indicating that uncertainty in candidate-set construction remains an important source of end-to-end errors. These findings further suggest that RexDrug’s reasoning-driven generative paradigm is more robust than surface-pattern matching for modeling complex pharmacological semantics. (5) We further extend RexDrug to jointly output all drug entities and relation tuples. Despite the additional output requirements, RexDrug continues to outperform all baselines under the same configuration and achieves an F1 score of 95.1\% on the NER subtask, demonstrating its adaptability to different structured-output requirements. Complete results and analyses are provided in Supplementary Section S4.

\begin{figure*}[t]
\centering
    \includegraphics[width=\textwidth]{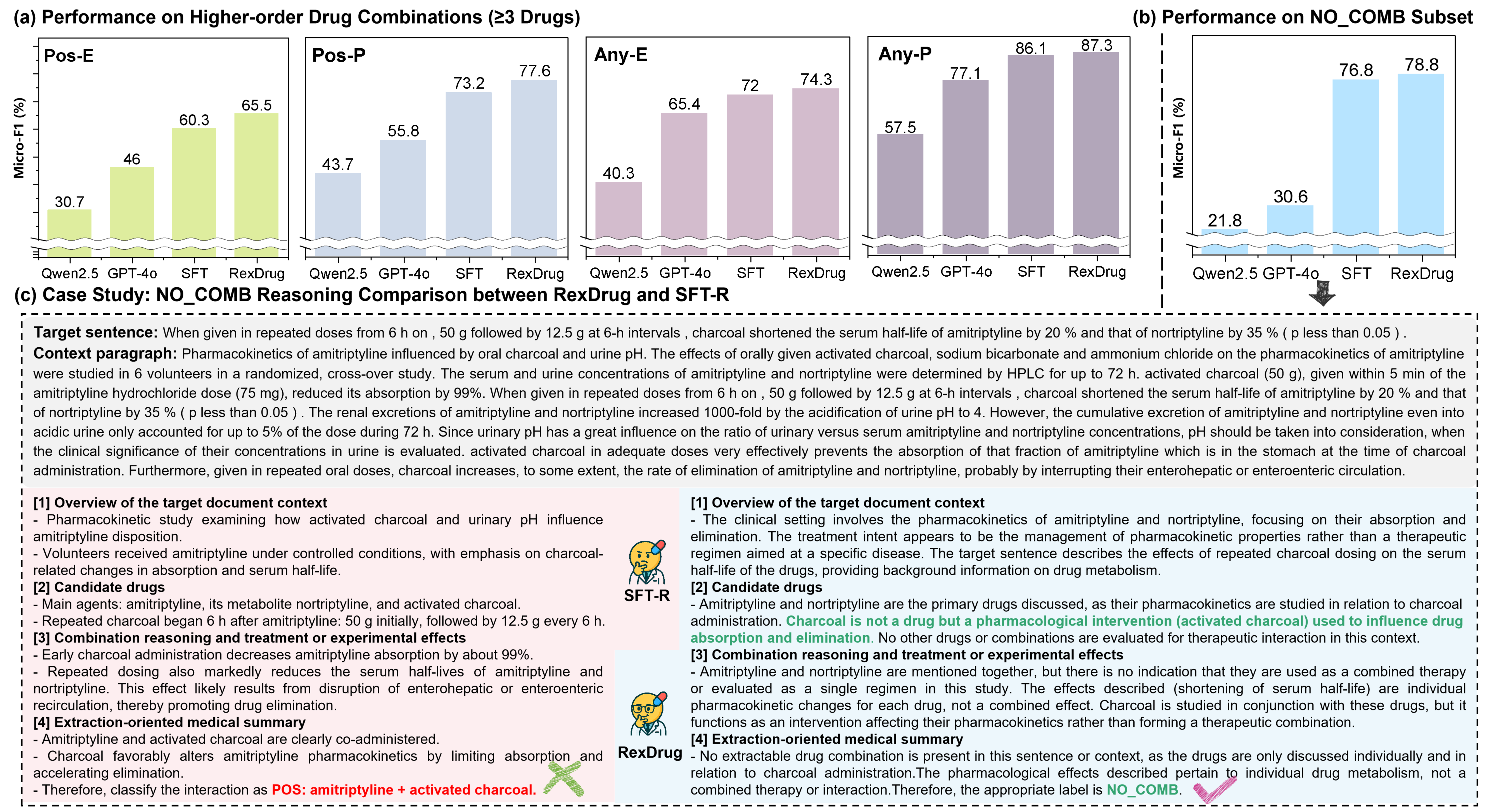}
    \caption{Fine-grained robustness analysis of RexDrug. (a) Higher-order drug combinations ($\geq$3 drugs): performance on Pos-E, Pos-P, Any-E, and Any-P. "Pos" and "Any" indicate metrics for positive combinations and all combinations, respectively; "-E" and "-P" refer to exact match and partial match scores. (b) NO\_COMB negatives: discrimination performance measured by micro-F1 on the NO\_COMB subset. (c) NO\_COMB case study: qualitative comparison of SFT-R (SFT-R denotes the supervised fine-tuned model that is required to generate reasoning traces.) and RexDrug on a representative NO\_COMB example.}
    \label{Fig3}
\end{figure*}

\begin{figure}[t]
\centering
    \includegraphics[width=\columnwidth]{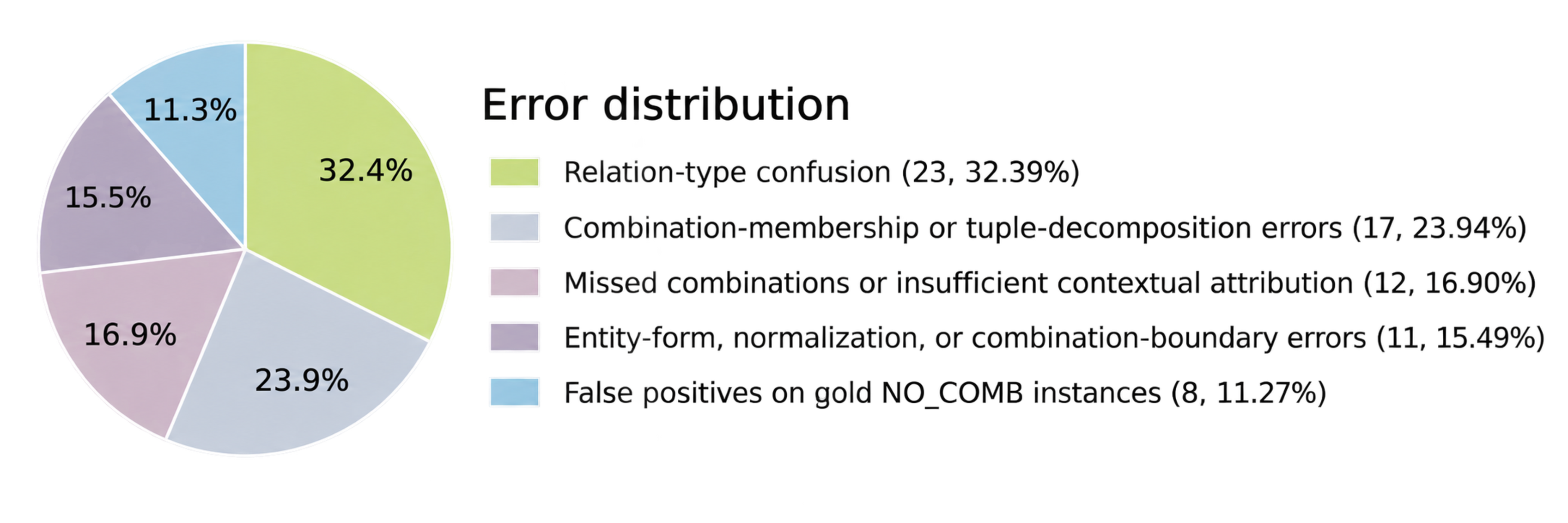}
    \caption{Error category distribution of RexDrug-LLaMA on the DrugComb test set. The analysis summarizes 71 non-exact predictions among 272 test instances.}
    \label{Fig5}
\end{figure}
\subsection{Evaluation on Fine-grained Robustness}

To further probe model behavior at specific logical boundaries, we conduct a fine-grained analysis on two challenging subsets, namely NO\_COMB (negative samples) and higher-order \textit{n}-ary combinations (involving three or more drugs). First, distinguishing true pharmacological interactions from non-combinations is a central challenge in DCE. As shown in Fig. \ref{Fig3}(b), general-purpose LLMs (e.g., GPT-4o and Qwen2.5) exhibit substantial over-extraction on the NO\_COMB subset. In contrast, RexDrug-Qwen, equipped with cognitive priming and logic alignment, better internalizes the criteria for non-combination cases and effectively suppresses false positives under complex contexts. As illustrated in the example in Fig. \ref{Fig3}(c), RexDrug correctly recognizes that charcoal is not a conventional "drug" entity but rather a pharmacological intervention, through regimen-level reasoning, and concludes that no valid drug combination is present in the text. Second, as shown in Fig. \ref{Fig3}(a), RexDrug achieves a clear improvement over its SFT baseline on higher-order \textit{n}-ary combinations, suggesting that as interaction complexity increases, RexDrug can more accurately disentangle the semantic characteristics of higher-order interactions and more reliably infer relations within complex regimens. 

Despite these gains, RexDrug retains several failure modes. An instance-level analysis of all 272 DrugComb test instances using RexDrug-LLaMA identified 201 exact predictions and 71 instances with at least one error. As shown in Fig. \ref{Fig5}, relation-type confusion was most frequent, particularly between COMB and POS, when co-administration, standard-regimen descriptions, or document-level outcomes were mistaken for combination-specific benefits. Combination-membership and tuple-decomposition errors also occurred, often because drugs from distinct regimens were conflated. Among the 97 gold NO\_COMB instances, 8 were incorrectly predicted as positive cases, primarily due to resistance-related contexts and misleading lexical cues such as "combined with". Qualitative examples are provided in Supplementary Section S6.

\subsection{Performance on the External DDI13 Dataset}
To assess the applicability of RexDrug, we evaluate its performance on the binary DDI13 dataset \cite{herrero2013ddi} under both relation extraction and relation classification settings. Accordingly, we slightly adapt the metric-based reward to match the DDI13 evaluation protocol: instead of the Exact/Partial matching used for \textit{n}-ary DCE, we compute the Combination Metric Reward using the micro-averaged F1 score over DDI relation types. This adjustment keeps the RL objective aligned with the dataset-specific metric while leaving the other reward components unchanged. In addition to GPT-based models, we include the following baselines: \textbf{Relation Extraction: }TP-DDI \cite{zaikis2021tp} uses a BioBERT-based pipeline for sequential entity and relation prediction; MRC4BioER \cite{sun2022mrc4bioer} reformulates the task as machine reading comprehension using query-style prompts; SPBRE \cite{yang2023spbere} performs span enumeration with type-marked inputs for relation extraction. \textbf{Relation Classification: }AW-BLSTMs \cite{mostafapour2019attention} employs hierarchical BiLSTMs with dual attention; BERTKG-DDI \cite{mondal2021bertkg} enhances BioBERT with knowledge graph features; SubGE-DDI \cite{shi2024subge} fuses PubMedBERT embeddings with drug subgraph graph convolutional network (GCN) representations.

\begin{table}[t]
\caption{
Performance comparison on the DDI13 dataset. “RE” denotes relation extraction without gold entity information, whereas “RC” denotes relation classification with gold entities. RexDrug results are reported as mean ± standard deviation over five independent runs.
}
\centering
\begin{tabular}{llccc}
\toprule
\textbf{Task} & \textbf{Method} & \textbf{P (\%)} & \textbf{R (\%)} & \textbf{F1 (\%)} \\
\midrule
\multirow{6}{*}{\textbf{RE}} 
    & GPT-3.5         & 35.6 & 26.9 & 30.7 \\
    & GPT-4o          & 53.9 & 54.4 & 54.2  \\
    & TP-DDI          & \underline{86.4} & \textbf{78.8} & \underline{82.4} \\
    & MRC4BioER       & 76.9 & 74.6 & 75.7 \\
    & SPBRE           & 78.5 & 79.7 & 79.2 \\
    & RexDrug-Qwen & \textbf{87.0{$\pm$}1.3} & \underline{78.9$\pm$1.6} & \textbf{82.7$\pm$0.6} \\
\midrule
\multirow{6}{*}{\textbf{RC}}
    & GPT-3.5         & 56.0 & 39.9 & 46.6 \\
    & GPT-4o          & 58.9 & 68.1 & 63.2 \\
    & AW-BLSTMs       & 80.0 & 77.0 & 78.5 \\
    & BERTKG-DDI      & --    & --    & \underline{84.0} \\
    & SubGE-DDI       & \underline{85.0} & \underline{82.8} & 83.9 \\
    & RexDrug-Qwen & \textbf{88.4{$\pm$}1.9} & \textbf{86.9$\pm$2.4} & \textbf{87.6$\pm$0.4} \\
\bottomrule
\end{tabular}
\label{tab:ddi-comparison}
\end{table}

As shown in Table~\ref{tab:ddi-comparison}, across five independent runs, RexDrug achieves an F1 score of 82.7\% in the relation extraction setting, which further increases to 87.6\% in the relation classification setting, surpassing the best baseline by 3.6\%. This improvement demonstrates that, even in structurally simpler tasks, RexDrug effectively captures subtle interaction signals and avoids overfitting to superficial co-occurrence patterns. Instead, it aligns its outputs with underlying medical logic. Overall, these results show that RexDrug not only improves performance in \textit{n}-ary relation extraction but also generalizes well across different relation extraction datasets and task settings, offering a unified and robust solution for biomedical relation extraction. To further illustrate the model's behavior beyond aggregate metrics, Supplementary Fig. S2 presents a representative DDI13 case.

\begin{table}[t]
\centering
\begin{minipage}{\columnwidth}
\caption{
Ablation study results on the DrugComb dataset evaluating the impact of different reward function components in RexDrug, based on the Qwen2.5-7B-Instruct model.
}
\centering
\setlength{\tabcolsep}{4pt}
\begin{tabular}{lcccc}
\toprule
\textbf{Category} & \textbf{Pos-E} & \textbf{Pos-P} & \textbf{Any-E} & \textbf{Any-P} \\
\midrule
RexDrug & \textbf{74.2$\pm$0.4} & \textbf{76.9$\pm$0.8} & \textbf{81.3$\pm$0.6} & \textbf{84.1$\pm$0.5} \\
w/o $r_{format}$ & 60.6$\pm$2.4 & 65.2$\pm$1.6 & 64.3$\pm$1.8 & 69.5$\pm$0.9 \\
w/o $r_{cover}$ & 68.2$\pm$1.2 & 72.7$\pm$0.9 & 71.1$\pm$0.8 & 75.8$\pm$0.7 \\
w/o $r_{metric}$ & 66.9$\pm$2.9 & 72.9$\pm$3.4 & 71.8$\pm$1.7 & 78.1$\pm$1.4 \\
\bottomrule
\end{tabular}
\label{tab:ablation-compact2}
\end{minipage}
\end{table}

\subsection{Ablation Study}
We conduct ablation experiments on the reward components within the full RexDrug framework. As shown in Table \ref{tab:ablation-compact2}: (1) Removing the structural format reward (w/o $r_{format}$) leads to the largest performance drop, with the POS-E F1 decreasing by 13.6 F1 points. Although instruction-following is partially learned during the warm-up, generative models still produce outputs that violate format constraints. This highlights the critical role of format guidance in maintaining output structure and ensuring downstream usability. (2) Removing the drug combination metric reward (w/o $r_{metric}$) also causes a substantial performance degradation, as it eliminates an important optimization signal for guiding the model to recover multi-drug combination structures. (3) Removing the combination coverage reward with the empty output penalty (w/o $r_{cover}$) leads to a performance drop. Under this setting, the model tends to over-predict empty outputs due to the high proportion of NO\_COMB instances in the dataset, thereby hurting recall.

\section{Discussion}

\subsection{Is Multi-Agent reasoning trace Generation Better than Single-Model Generation?}
To support the supervised warm-up (SFT) stage of RexDrug, we propose a multi-agent collaborative strategy for reasoning-trace generation. Compared with single-model self-generation, this strategy incurs higher computational and API costs, which raises a natural question: does iterative multi-agent collaboration indeed produce higher-quality data? To answer this, we randomly sampled 200 instances and collected the corresponding reasoning traces. These traces were independently evaluated by a graduate student with a background in medicine and pharmacological research. The evaluator found that the multi-agent-generated traces received mean medical and logical-accuracy scores of 4.70 and 4.68, respectively (rated on a 0–5 scale), suggesting that the data quality is sufficient for both training and interpretation. The detailed evaluation criteria and illustrative scoring examples are provided in Supplementary Section S5.

\begin{figure*}[t]
\centering
    \includegraphics[width=\textwidth]{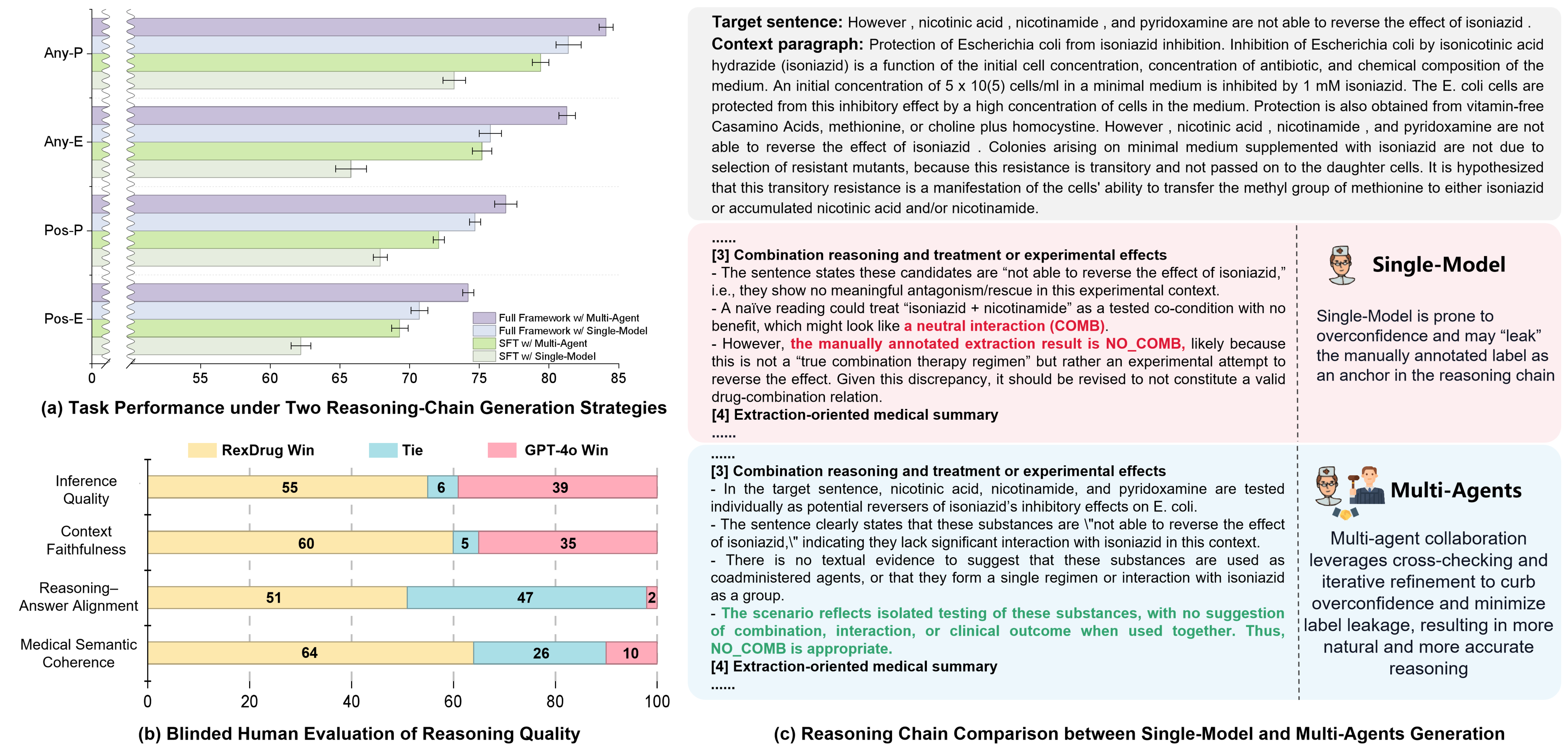}
    \caption{Evaluation and qualitative analysis of the RexDrug reasoning framework. (a) Task performance under two reasoning strategies. "Full Framework" denotes the complete RexDrug training pipeline. (b) Blinded human evaluation of 100 test instances by two graduate raters with medical and pharmacological training; win, tie, and loss counts are determined by comparing model scores averaged across raters for each instance and dimension. (c) Case study contrasting reasoning traces generated under the single-model and multi-agent settings.}
    \label{Fig4}
\end{figure*}

We further examine the impact of reasoning-trace quality by comparing two training settings, SFT-only and SFT followed by RL. As shown in Fig. \ref{Fig4}(a), reasoning traces produced by a single model lead to systematic performance degradation across all metrics, suggesting that the absence of a review-and-correction mechanism introduces substantial noise. In contrast, the more consistent and lower-noise reasoning traces generated by the multi-agent strategy provide more reliable supervision and translate into improved extraction performance.

\subsection{How Usable Are RexDrug-Generated Reasoning Traces?}

\begin{table}[t]
\centering
\caption{Evaluation of reasoning quality on DrugComb using ROSCOE metrics. SFT-R denotes supervised fine-tuning on distilled reasoning traces, whereas RexDrug further applies GRPO; both use the Qwen2.5-7B-Instruct backbone.}
\label{tab:roscoe}
\setlength{\tabcolsep}{2pt}
\renewcommand{\arraystretch}{1.05}
\begin{tabular}{
@{}c@{\hspace{2pt}}
>{\centering\arraybackslash}p{2.8cm}@{\hspace{2pt}}
c@{\hspace{2pt}}c@{\hspace{2pt}}c@{}
}
\toprule
\textbf{Category} & \textbf{Metric} & \textbf{GPT-4o} & \textbf{SFT-R} & \textbf{RexDrug} \\
\midrule
\multirow{2}{*}{SA}
  & Faithfulness-Step  & \textbf{0.9010} & 0.8789 & \underline{0.8951} \\
  & Faithfulness-Token & \underline{0.9180} & 0.9135 & \textbf{0.9294} \\
\midrule
\multirow{2}{*}{SS}
  & Info-Step          & \underline{0.8977} & 0.8848 & \textbf{0.8991} \\
  & Repetition-Token   & \textbf{0.0556} & 0.0443 & \underline{0.0505} \\
\midrule
\multirow{2}{*}{LI}
  & Source Consistency & \textbf{0.8901} & 0.7816 & \underline{0.8069} \\
  & Self-Consistency   & \textbf{0.9135} & 0.7894 & \underline{0.7948} \\
\midrule
\multirow{3}{*}{LC}
  & Perplexity-Step    & \underline{0.0157} & 0.0117 & \textbf{0.0158} \\
  & Perplexity-Chain   & 0.0865 & \textbf{0.1170} & \underline{0.1139} \\
  & Grammar            & 0.9479 & \underline{0.9678} & \textbf{0.9697} \\
\bottomrule
\end{tabular}
\end{table}

We assess reasoning quality using ROSCOE \cite{golovneva2023roscoe}, which measures Semantic Alignment (SA), Semantic Similarity (SS), Logical Inference (LI), and Linguistic Coherence (LC) on a 0-1 scale. For multi-step biomedical reasoning, we modify LI by averaging contradiction scores across steps rather than taking the maximum, reducing sensitivity to isolated contradictions. As shown in Table \ref{tab:roscoe}, reasoning-supervised SFT with Qwen2.5-7B underperforms GPT-4o on most metrics. Applying GRPO to the same backbone improves most metrics and matches or exceeds GPT-4o on several, indicating that reward alignment enhances evidence grounding and linguistic coherence without increasing model scale.

We further conducted a blinded evaluation of 100 test instances with two graduate raters trained in medicine and pharmacology. Each rater reviewed the source text and anonymized reasoning outputs and assigned 0-5 scores for Context Faithfulness, Reasoning-Answer Alignment, Medical Semantic Consistency, and Inference Quality. Rater-averaged scores determined wins, ties, and losses, with inter-rater agreement reported in Supplementary Section S5. As shown in Fig. \ref{Fig4}(b), RexDrug achieved higher ratings more frequently, particularly for Context Faithfulness and Medical Semantic Consistency. GPT-4o occasionally introduced assumptions unsupported by the provided evidence, whereas RexDrug demonstrated stronger evidence grounding and closer alignment between reasoning trajectories and extraction outputs.

These results indicate that RexDrug produces accurate extractions with inspectable, evidence-grounded reasoning that may support knowledge-base construction, systematic reviews, and predictive models requiring structured drug inputs. We illustrate this technical connectivity through a retrospective workflow with MultiComb \cite{monem2024multi}, which predicts synergy and sensitivity for predefined drug-pair and cell-line inputs. After manual verification, drug-name normalization, and cell-line matching, RexDrug extractions are passed to MultiComb for context-specific scoring. A complete example is provided in Supplementary Section S7, with two additional cases available in the code repository.

\section{Conclusion}
This study proposes RexDrug, an end-to-end framework for Drug Combination Extraction (DCE) that aims to bridge the gap between automated information extraction and expert medical reasoning. To address the scarcity of annotated reasoning data in the biomedical domain, we develop a multi-agent collaboration mechanism to automatically synthesize high-quality reasoning-trace data for drug combinations. By combining medical cognitive priming with GRPO-based reinforcement learning, RexDrug can accurately extract drug combinations while generating structured and interpretable reasoning trajectories. Computational evaluations demonstrate that RexDrug achieves state-of-the-art performance on both the DrugComb and DDI13 datasets, and exhibits strong robustness in the quality of its generated reasoning traces. Future work will integrate external biomedical knowledge and extend RexDrug to broader literature-mining settings, improving its utility for evidence curation and structured biomedical knowledge construction.

\section*{Data availability}
The datasets used in this study are publicly available. The DrugComb dataset (drug-combo-extraction) \cite{tiktinsky2022dataset} is available via Hugging Face at \url{https://huggingface.co/datasets/allenai/drug-combo-extraction}. The DDI13 dataset is available from the SubGE-DDI \cite{herrero2013ddi} repository at \url{https://github.com/syy528000/SubGE-DDI}. The preprocessing scripts and all experimental code for RexDrug are available at \url{https://github.com/DUTIR-BioNLP/RexDrug}.

\bibliographystyle{unsrtnat}
\bibliography{references}  %%% Uncomment this line and comment out the ``thebibliography'' section below to use the external .bib file (using bibtex) .





% \end{thebibliography}
\clearpage
\appendix
\renewcommand{\thesection}{\Alph{section}}          % 主节编号 A, B, C...
\renewcommand{\thesubsection}{\thesection.\arabic{subsection}}  % 子节编号 C.1, C.2...

\section{Gold-Label-Anchored versus Label-Blind Trace Generation}\label{sec11}

Gold labels in our multi-agent pipeline constrain entity membership and relation types, thereby reducing task drift during trace generation. However, label access may induce answer-conditioned rationalization, even though the prompts require forward inference and prohibit wording that presupposes the answer. We therefore compared label-conditioned distillation with label-blind generation followed by rejection sampling.

For each of the 1362 DrugComb training instances, GPT-4o generated four candidate traces from the target sentence and its context without access to the gold relation. Gold labels were applied only after generation to retain candidates with correct final predictions. This procedure yielded accepted traces for 802 instances (58.88\%), compared with 1098 (80.62\%) under label-conditioned multi-agent distillation. Thus, label conditioning increased the efficiency of accepted-trace generation. The resulting larger training set also achieved higher overall scores, although this comparison confounds trace source with training-set size and composition.

To control for these factors, we trained separate Qwen2.5-7B-Instruct models on the 658 instances retained by both pipelines. Performance differences were small and inconsistent across metrics (Supplementary Table \ref{tab:trace_source}), providing no evidence of a uniform downstream advantage for label-conditioned traces on matched data.

Finally, we assessed whether performance reflected memorization of drug combinations or pairs shared across the training and test sets. The overlap rates were 11.50\% for DrugComb and 3.93\% for DDI13. Removing these test instances produced only small, non-systematic changes on DrugComb (Supplementary Table \ref{tab:overlap_analysis}), while DDI13 micro-F1 changed from $82.7\pm0.6$ to $82.9\pm0.4$. These results indicate that performance was not primarily driven by cross-split drug repetition.

\begin{table*}[htbp]
\caption{DrugComb results for label-blind rejection sampling (Label-blind) and label-conditioned multi-agent distillation (Label-cond.) using Qwen2.5-7B-Instruct. Values are mean$\pm$standard deviation over five runs; bold indicates the best result within each setting.}
\label{tab:trace_source}
\centering
\setlength{\tabcolsep}{6pt}
\renewcommand{\arraystretch}{1}
\scriptsize
\begin{tabular}{lccccc}
\toprule
\textbf{Trace Source} & \textbf{\#Train} & \textbf{Pos-E} & \textbf{Pos-P} & \textbf{Any-E} & \textbf{Any-P} \\
\midrule
\multicolumn{6}{l}{\textit{All retained instances}} \\
Label-blind
& 802
& 69.8$\pm$0.8
& 72.9$\pm$0.2
& 77.5$\pm$0.7
& 81.1$\pm$0.4 \\
Label-cond.
& 1098
& \textbf{74.2$\pm$0.4}
& \textbf{76.9$\pm$0.8}
& \textbf{81.3$\pm$0.6}
& \textbf{84.1$\pm$0.5} \\
\addlinespace[2pt]
\multicolumn{6}{l}{\textit{Matched instances}} \\
Label-blind
& 658
& \textbf{70.10$\pm$0.3}
& 72.46$\pm$0.2
& \textbf{77.07$\pm$0.9}
& \textbf{80.30$\pm$0.5} \\
Label-cond.
& 658
& 69.65$\pm$0.6
& \textbf{73.76$\pm$0.4}
& 76.62$\pm$0.7
& 80.03$\pm$0.7 \\
\bottomrule
\end{tabular}
\end{table*}

\begin{table}[t]
\caption{DrugComb performance on the original test set and after removing instances whose drug combinations appeared in training. Values are mean$\pm$standard deviation over five runs.}
\label{tab:overlap_analysis}
\centering
\setlength{\tabcolsep}{6pt}
\renewcommand{\arraystretch}{1}
\scriptsize
\begin{tabular}{lcccc}
\toprule
\textbf{Test Set} & \textbf{Pos-E} & \textbf{Pos-P} & \textbf{Any-E} & \textbf{Any-P} \\
\midrule
Original
& 74.2$\pm$0.4
& 76.9$\pm$0.8
& 81.3$\pm$0.6
& 84.1$\pm$0.5 \\
Filtered
& 73.8$\pm$0.3
& 77.3$\pm$0.5
& 80.6$\pm$0.3
& 83.8$\pm$0.6 \\
\bottomrule
\end{tabular}
\end{table}

\section{Failure Analysis and Selection Bias of Reasoning-Trace Filtering}

After three generation-review-feedback rounds, 264 of 1362 DrugComb training instances were excluded. Manual audit assigned one primary cause to each instance: explicit reference to the provided annotation or answer-anchored post hoc reasoning (223, 84.5\%); unsupported efficacy, synergy, or toxicity claims (14, 5.3\%); inconsistency between the extraction and reference annotation (13, 4.9\%); misclassification of separate experimental arms, alternative regimens, or sequential therapies as combinations (7, 2.7\%); and combination-boundary errors or omitted drugs (5, 1.9\%). Of the remaining two instances, one involved a conflict between the textual evidence and annotation, and the other exceeded the reasoning-length limit. Formatting alone caused no exclusions because such errors were corrected through iterative feedback.

To assess selection bias, we compared retained and excluded instances by input length, candidate-drug count, relation count, combination arity, and relation category. Exclusion showed no monotonic increase with input length, candidate-set size or relation count. Among instances with annotated combinations, excluding NO\_COMB cases, the rates were 25.5\% for binary, 9.1\% for ternary, and 5.6\% for combinations involving four or more drugs. By relation category, the rates were 18.6\% for POS, 17.8\% for COMB, 18.0\% for NO\_COMB, and 29.2\% for NEG. Note that, DrugComb contains POS, COMB, and NEG relation annotations together with no-combination instances; following the evaluation protocol, COMB and NEG are merged into OTHER. Thus, filtering did not preferentially remove structurally complex instances, although the higher NEG exclusion rate suggests that negative-effect reasoning may be underrepresented in the SFT trace set.

\section{Experimental Details}\label{sec33}

\begin{table}[!htbp]
\caption{Detailed statistics for the DrugComb dataset}
\centering
\begin{tabular}{lrrr}
\toprule
\textbf{Statistic}                & \textbf{Train} & \textbf{Test} & \textbf{Total} \\
\toprule
\multicolumn{4}{l}{\textit{Document-level}} \\
\midrule
No relation                      & 494   & 97    & 591  \\
One relation                     & 731   & 146   & 877  \\
More than one relation           & 137   & 29    & 166  \\
\midrule
\multicolumn{4}{l}{\textit{Combination-level}} \\
\midrule
Binary                           & 745   & 155   & 900  \\
3-ary                            & 191   & 35    & 226  \\
4-ary                            & 57    & 12    & 69   \\
5-ary or more                    & 46    & 7     & 53   \\
\bottomrule
\end{tabular}
\label{tab:DrugComb}
\end{table}

\begin{table}[!htbp]
\caption{Detailed statistics for the DDI13 dataset. \textbf{\# Sentences} denotes the number of sentence-level data instances. \textbf{Adv}, \textbf{Eff}, \textbf{Int}, and \textbf{Mec} indicate the number of positive pairs for Advice, Effect, Interaction, and Mechanism relations, respectively.}
\centering
\begin{tabular}{lrrrrr}
\toprule
        & \textbf{\# Sentences} & \textbf{Adv} & \textbf{Eff} & \textbf{Int} & \textbf{Mec} \\[1pt]
\midrule
Train   & 3135    & 697    & 1347   & 157    & 1193 \\[2pt]
Test    &  697    & 189    & 297    & 79     & 285  \\
\bottomrule
\end{tabular}
\label{tab:ddi2013}
\end{table}

We conducted computational evaluations on the DrugComb and DDI13 datasets. Detailed information about these datasets is provided in Supplementary Table \ref{tab:DrugComb} and Table \ref{tab:ddi2013}.

For both datasets, GPT-4o (\texttt{gpt-4o-2024-11-20}) served as the Medical Reasoning Analyst and GPT-5.1 (\texttt{gpt-5.1-2025-11-13}) as the Medical Expert Reviewer. The latter is a publicly accessible API model with a 400000-token context window and a 128000-token maximum output. After review, the pipeline retained 1098 of 1362 DrugComb instances and 2907 of 3135 DDI13 instances. Trace construction required 11678 API calls, averaging 2.60 per training instance, and consumed 27809499 tokens. Monetary costs depend on provider pricing at the time of use. These API costs were incurred only during offline trace construction; inference uses the fine-tuned backbone without further GPT-4o or GPT-5.1 calls. Model snapshots and message content were fixed, whereas temperature, top\_p, max\_tokens, and seed were left unspecified and followed the API defaults.

We fine-tuned LLaMA3.1-8B-Instruct and Qwen2.5-7B-Instruct using LoRA (rank 64, $\alpha=128$) in both SFT and GRPO. SFT used a learning rate of $1\times10^{-6}$ for 10 epochs. GRPO used the same learning rate for 20 epochs, with a maximum input length of 2,048 tokens, temperature 0.4, and group size 8. Each trainable configuration was independently run with five random seeds (42, 123, 2025, 3407, and 6666), with results reported as mean$\pm$standard deviation. Training was conducted on a single NVIDIA A100 GPU using PyTorch 2.8.0 and the Swift framework. Across the two backbones, the maximum SFT and GRPO training times were 1.95 and 44.30 GPU-hours for DrugComb and 3.65 and 127.20 GPU-hours for DDI13, respectively.

We evaluated reward-weight sensitivity using five Qwen2.5-7B-Instruct configurations on DrugComb under the same protocol and random seeds. The selected configuration yielded the best result on three of four metrics (Supplementary Table \ref{tab:reward_sensitivity}). Increasing the metric-reward weight to 0.8 improved Pos-E alone, whereas format- or coverage-dominant settings performed worse overall, supporting metric reward as the primary signal and the other rewards as auxiliary constraints.

\begin{figure*}[h]
\centering
\includegraphics[width=\textwidth]{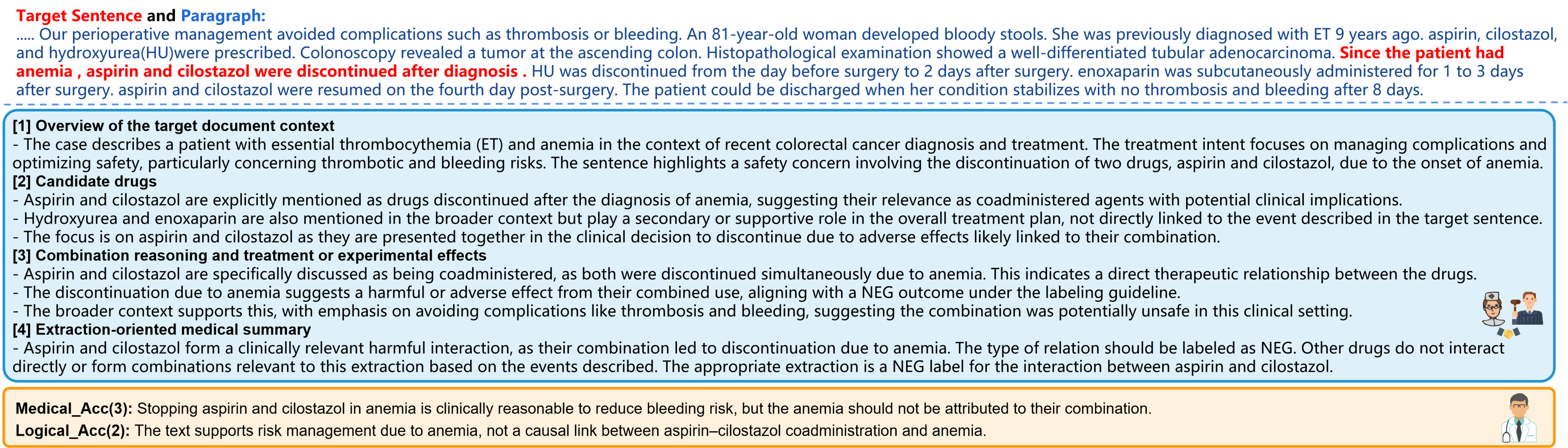}
\caption{Representative penalized example illustrating that the reasoning chain incorrectly attributes anemia to the aspirin–cilostazol combination and lacks sufficient evidence-to-conclusion bridging, resulting in reduced medical and logical accuracy scores.}
\label{a-mutiagent-error}
\end{figure*}

\section{Extended Evaluation of RexDrug on Entity Recognition and Relation Extraction}\label{sec44}

\begin{table}[t]
\caption{Reward-weight sensitivity on DrugComb using Qwen2.5-7B-Instruct. Weights are ordered as $(r_{format}, r_{comb\_cover}, r_{comb\_metric})$.}
\label{tab:reward_sensitivity}
\centering
\setlength{\tabcolsep}{6pt}
\renewcommand{\arraystretch}{1}
\scriptsize
\begin{tabular}{lcccc}
\toprule
\textbf{Weights} & \textbf{Pos-E} & \textbf{Pos-P} & \textbf{Any-E} & \textbf{Any-P} \\
\midrule
$(0.33, 0.33, 0.33)$
& 71.7$\pm$0.2 & 75.2$\pm$0.5 & 78.3$\pm$0.4 & 82.6$\pm$0.4 \\

$(0.8, 0.1, 0.1)$
& 71.8$\pm$0.7 & 74.7$\pm$0.5 & 78.1$\pm$0.6 & 81.6$\pm$0.8 \\

$(0.1, 0.8, 0.1)$
& 72.4$\pm$0.4 & 76.5$\pm$0.8 & 78.8$\pm$0.7 & 83.2$\pm$0.3 \\

$(0.1, 0.1, 0.8)$
& \textbf{75.5$\pm$0.2} & \underline{76.8$\pm$0.3} & \underline{80.4$\pm$0.9} & \underline{83.6$\pm$0.6} \\

$(0.2, 0.1, 0.7)$
& \underline{74.2$\pm$0.4} & \textbf{76.9$\pm$0.8} & \textbf{81.3$\pm$0.6} & \textbf{84.1$\pm$0.5} \\
\bottomrule
\end{tabular}
\end{table}

We extend the \textit{n}-ary drug-combination extraction (NDCE) task to require explicit output of all mentioned drug entities together with their combination relations, denoting this variant as RexDrug+. Whereas RexDrug learns entity recognition implicitly through relation supervision, RexDrug+ introduces explicit named entity recognition (NER) targets to evaluate their effect on both tasks.

RexDrug+ places entities within \texttt{@ner\#...\#ner@} and relations within \texttt{@re\#...\#re@}. Accordingly, RL replaces the original format reward with a structural reward that validates both tagged fields and their schemas, and adds an entity-level F1 reward. We use Seq2Rel+, which also jointly extracts entities and relations, as the baseline and apply the same output format to GPT-3.5 and GPT-4o. Relation-extraction metrics remain unchanged, while NER is evaluated using entity-level F1. As shown in Supplementary Table \ref{tab-NDCEPLUS}, RexDrug+ outperforms the joint-extraction baselines and achieves the highest NER F1 of 95.1. Relative to RexDrug, explicit entity supervision slightly improves Pos-P but reduces the other relation metrics, indicating a trade-off rather than a uniform gain. Nevertheless, RexDrug+ provides a practical option when both drug entities and their relations are required.

\begin{table}[t]
\caption{Joint entity and relation extraction on DrugComb. "+" denotes Seq2Rel and RexDrug variants with explicit entity output; their original variants use the standard NDCE setting.}
\label{tab-NDCEPLUS}
\centering
\setlength{\tabcolsep}{3pt}
\renewcommand{\arraystretch}{0.98}
\scriptsize
\begin{tabular}{lccccc}
\toprule
\textbf{Method} & \textbf{Pos-E} & \textbf{Pos-P} & \textbf{Any-E} & \textbf{Any-P} & \textbf{NER} \\
\midrule
Seq2Rel          & 66.7 & --   & 71.1 & --   & --   \\
Seq2Rel+         & 65.8 & --   & 74.0 & --   & 94.0 \\
\midrule
GPT-3.5 w/o reas.& 38.4 & 45.9 & 39.6 & 48.7 & 83.9 \\
GPT-3.5 w/ reas. & 33.9 & 36.9 & 34.9 & 40.3 & 84.8 \\
GPT-4o w/o reas. & 47.4 & 51.9 & 46.8 & 51.6 & 86.9 \\
GPT-4o w/ reas.  & 51.4 & 57.9 & 48.7 & 56.0 & 87.8 \\
\midrule
RexDrug (Qwen)   & \underline{74.2$\pm$0.4} & 76.9$\pm$0.8 & \underline{81.3$\pm$0.6} & 84.1$\pm$0.5 & -- \\
RexDrug+ (Qwen)  & 73.2$\pm$0.6 & \underline{77.9$\pm$1.2} & 77.4$\pm$0.8 & 82.1$\pm$1.7 & \underline{94.2$\pm$0.8} \\
\midrule
RexDrug (LLaMA)  & \textbf{74.6$\pm$0.7} & 77.1$\pm$0.5 & \textbf{81.4$\pm$0.5} & \textbf{85.3$\pm$0.6} & -- \\
RexDrug+ (LLaMA) & 73.8$\pm$0.4 & \textbf{78.1$\pm$0.9} & 77.3$\pm$1.2 & \underline{84.4$\pm$1.1} & \textbf{95.1$\pm$0.4} \\
\bottomrule
\end{tabular}
\end{table}

\section{Human Evaluation}\label{sec55}\label{sec66}

\noindent\textit{Multi-agent trace audit.}
We randomly sampled 200 reasoning traces from the DrugComb training set and asked a medical graduate student with pharmacology research experience to rate each trace on two 0--5 scales: medical accuracy, covering semantic interpretation and consistency with pharmacological or clinical knowledge; and logical accuracy, covering coherence, completeness, and evidential support. The mean scores were 4.70 and 4.68, respectively. Supplementary Fig. \ref{a-mutiagent-error} shows a penalized example involving contextual misinterpretation and insufficient inferential support.

\noindent\textit{Comparison with GPT-4o.}
We randomly sampled 100 test instances and generated reasoning traces from RexDrug and GPT-4o using identical inputs. Two medical graduate students with pharmacology backgrounds independently reviewed the original text and anonymized outputs, rating each trace from 0 to 5 on four dimensions: Context Faithfulness (fidelity to the source evidence), Reasoning--Answer Alignment (consistency between the reasoning and prediction), Medical Semantic Consistency (context-appropriate medical interpretation), and Inference Quality (coherence, completeness, and interpretability). For each instance and dimension, ratings were averaged across the two reviewers; the model with the higher mean was counted as the winner, and equal means as ties. Inter-rater agreement was assessed using linearly weighted Cohen's $\kappa$ with bootstrap 95\% confidence intervals and ranged from 0.704 to 0.986 (Supplementary Table \ref{tab:human_agreement}).

\begin{table}[t]
\caption{Inter-rater agreement in the RexDrug--GPT-4o comparison. Values are linearly weighted Cohen's $\kappa$ with bootstrap 95\% confidence intervals.}
\label{tab:human_agreement}
\centering
\setlength{\tabcolsep}{5pt}
\renewcommand{\arraystretch}{0.98}
\scriptsize
\begin{tabular}{lcc}
\toprule
\textbf{Dimension} & \textbf{Weighted $\kappa$} & \textbf{95\% CI} \\
\midrule
Context Faithfulness
& 0.704 & [0.612, 0.758] \\
Reasoning--Answer Alignment
& 0.986 & [0.972, 0.997] \\
Medical Semantic Consistency
& 0.741 & [0.648, 0.807] \\
Inference Quality
& 0.757 & [0.700, 0.804] \\
\bottomrule
\end{tabular}
\end{table}

Human evaluation complements automated metrics by directly assessing evidence fidelity, medical semantics, and explanatory coherence. RexDrug achieved higher mean scores than GPT-4o across all four dimensions, as shown in Fig. 4(b) of the main manuscript.

% \begin{figure*}[h]
% \centering
% \includegraphics[width=\textwidth]{wang_s3.eps}
% \caption{Case Study 1}
% \label{a2-case1}
% \end{figure*}

% \begin{figure*}[h]
% \centering
% \includegraphics[width=\textwidth]{wang_s4.eps}
% \caption{Case Study 2}
% \label{a2-case2}
% \end{figure*}

\section{Case Analysis}\label{sec77}\label{sec88}

\begin{figure*}[t]
\centering
\includegraphics[width=\textwidth]{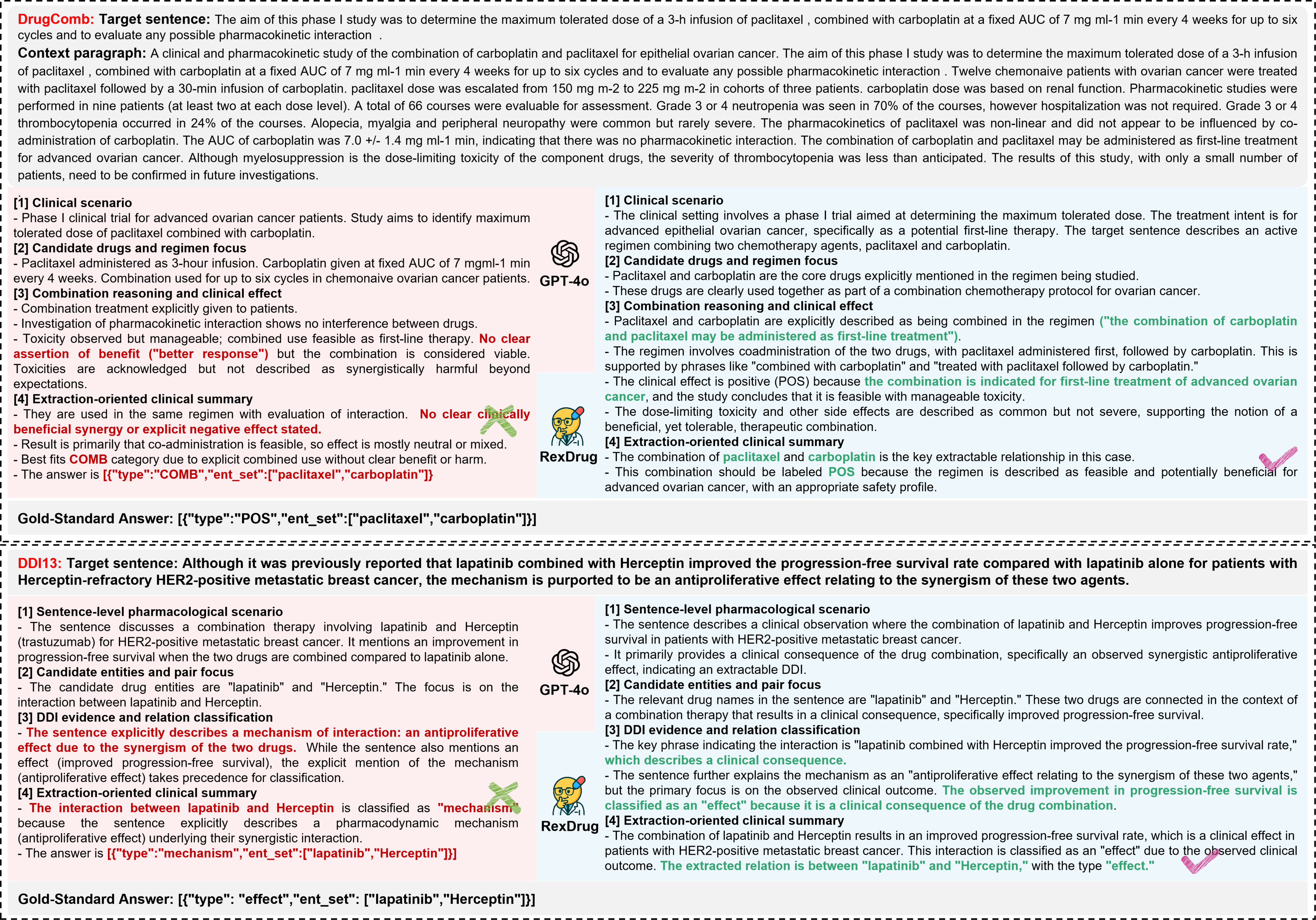}
\caption{Cross-dataset comparison of RexDrug and GPT-4o under identical instructions. The DrugComb case requires contextual evidence to distinguish POS from COMB, whereas the DDI13 case requires distinguishing an observed pharmacological effect from a mechanistic statement. Gold-standard outputs are shown.}
\label{fig:cross-dataset-cases}
\end{figure*}

Supplementary Fig. \ref{fig:cross-dataset-cases} presents one DrugComb and one DDI13 case. Because most task-specific baselines return only labels or tuples, GPT-4o serves as a reasoning-enabled qualitative comparator under identical task instructions. In the DrugComb case, the target sentence describes a phase I trial of paclitaxel combined with carboplatin, while evidence supporting POS is distributed across the context. GPT-4o identifies co-administration but predicts COMB because no explicit benefit is stated. RexDrug instead integrates the regimen's suitability for first-line treatment, absence of pharmacokinetic interaction, and manageable toxicity, producing the gold POS label. In the DDI13 case, the sentence reports improved progression-free survival with lapatinib and Herceptin before mentioning a purported antiproliferative mechanism. GPT-4o prioritizes the lexical cue "mechanism", whereas RexDrug anchors its classification to the observed outcome and correctly predicts \textit{effect}. These cases illustrate RexDrug's ability to integrate cross-sentence evidence and distinguish closely related relation types while providing an inspectable basis for each prediction. Five additional cases, including correct predictions and representative failures, are available in the code repository.

\section{Illustrative RexDrug--MultiComb Workflow}\label{sec99}

RexDrug extracts evidence-supported drug combinations and relation types from biomedical literature, whereas MultiComb predicts synergy and sensitivity for predefined drug-pair and cell-line inputs using molecular structures and gene expression. To examine their technical connectivity, we retrained MultiComb on the O'Neil dataset under the reported five-fold protocol and selected drug-pair and cell-line triplets represented in both O'Neil and relevant PubMed articles. RexDrug processed only the literature text; its extractions were manually verified, normalized, and matched to the corresponding O'Neil records before MultiComb scoring. This retrospective workflow does not constitute prospective combination discovery.

\begin{figure*}[t]
\centering
\includegraphics[width=\textwidth]{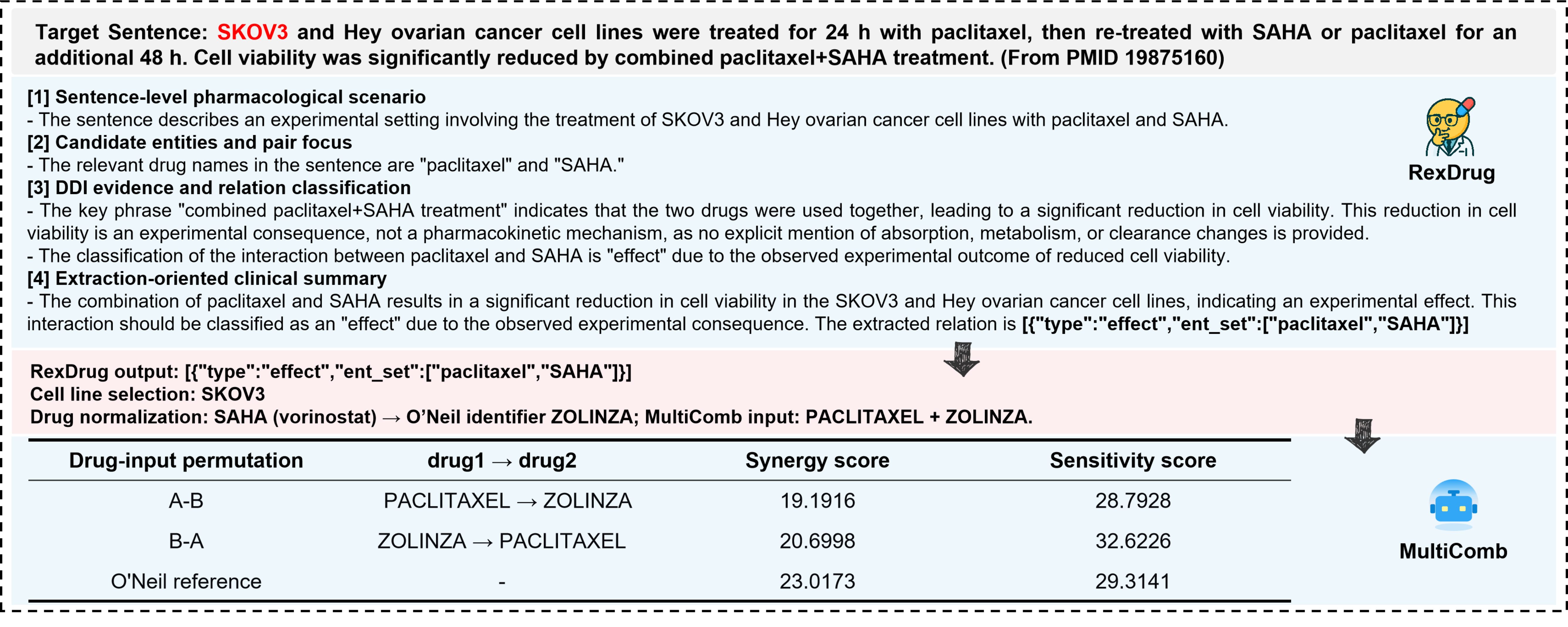}
\caption{Illustrative RexDrug--MultiComb workflow. RexDrug extracts an \textit{effect} relation between paclitaxel and SAHA in SKOV3 cells. After SAHA is mapped to vorinostat and the O'Neil identifier ZOLINZA, the normalized drug pair and cell line are passed to MultiComb for synergy and sensitivity prediction. Both input orders and the matched O'Neil reference values are shown.}
\label{a2-case3}
\end{figure*}

Fig.~\ref{a2-case3} presents paclitaxel and SAHA in SKOV3 ovarian cancer cells. RexDrug extracted an evidence-supported \textit{effect} relation from literature reporting reduced viability after combined exposure. SAHA was normalized to vorinostat and mapped to the O'Neil identifier ZOLINZA. Both drug-input orders produced positive synergy estimates close to the O'Neil reference; the sensitivity estimate was similarly close under the A-B order but higher under B-A. This case illustrates how literature-derived relations can be linked to cell-line-specific scoring after entity normalization and dataset matching. Two additional cases evaluated under the same protocol are available in the code repository\footnote{\url{https://github.com/DUTIR-BioNLP/RexDrug}}.

\end{document}